\documentclass[letterpaper, 10 pt, conference]{ieeeconf}  

\IEEEoverridecommandlockouts                              

\usepackage{graphicx} 
\usepackage{epsfig} 
\usepackage{times} 
\usepackage{amsmath} 
\usepackage{amssymb}  
\usepackage{biblatex}
\usepackage{booktabs}
\usepackage{placeins}
\usepackage{multicol}
\usepackage{multirow}
\usepackage{graphicx}
\usepackage{textcomp}
\usepackage[dvipsnames]{xcolor}
\usepackage{textcase}
\usepackage[tablename=TABLE]{caption}
\let\labelindent\relax
\usepackage{enumitem}
\usepackage{diagbox}
\usepackage{caption}
\usepackage{float}
\usepackage{hyperref}
\usepackage{gensymb}
\newcommand{\tabincell}[2]{\begin{tabular}{@{}#1@{}}#2\end{tabular}}

\title{\LARGE \bf
Learn to Differ: Sim2Real Small Defection Segmentation Network
}

\author{Zexi Chen, Zheyuan Huang, Yunkai Wang, Xuecheng Xu, Yue Wang, Rong Xiong
\thanks{All authors are with the State Key Laboratory of Industrial Control Technology and Institute of Cyber-Systems and Control, Zhejiang University, Zhejiang, China. Yue Wang is the corresponding author {\tt\small wangyue@iipc.zju.edu.cn}.}}

\begin{document}

\maketitle
\thispagestyle{empty}
\pagestyle{empty}


\begin{abstract} Recent studies on deep-learning-based small defection segmentation approaches are trained in specific settings and tend to be limited by fixed context. Throughout the training, the network inevitably learns the representation of the background of the training data before figuring out the defection. They underperform in the inference stage once the context changed and can only be solved by training in every new settings. This eventually leads to the limitation in practical robotic applications where contexts keep varying. To cope with this, instead of training a network context by context and hoping it to generalize, why not stop misleading it with any limited context and start training it with pure simulation? In this paper, we propose the network \textit{SSDS} that learns a way of distinguishing small defections between two images regardless of the context, so that the network can be trained once for all. A small defection detection layer utilizing the pose sensitivity of phase correlation between images is introduced and is followed by an outlier masking layer. The network is trained on randomly generated simulated data with simple shapes and is generalized across the real world. Finally, \textit{SSDS} is validated on real-world collected data and demonstrates the ability that even when trained in cheap simulation, \textit{SSDS} can still find small defections in the real world showing the effectiveness and its potential for practical applications. {\small \href{https://github.com/jessychen1016/SSDS}{\texttt{Code is available here}}}
\end{abstract}

\IEEEpeerreviewmaketitle

\section{Introduction}
\label{sec:introduction}
Imagery-based small defection segmentation (DS) always play important roles in modern assembly line. It is attractive due to its sensory convenience especially under the prosperity of high-resolution camera equipment. Small defections can be either of known types or random objects, considering which derives two fashions of DS. To deal with the former one, vision-based classifier and generator\cite{akcay2018ganomaly,andrews2016transfer,napoletano2018anomaly, schlegl2017unsupervised} emerged but failed in dealing with those unseen random objects. To recognize the latter, defected and defection-free templates are compared\cite{ding2019tdd,gaidhane2018efficient,moganti1995automatic, wang2016framework}, which also leaves us with a great many problems to be solved.

    \begin{figure}[t]
        \includegraphics[width=\linewidth]{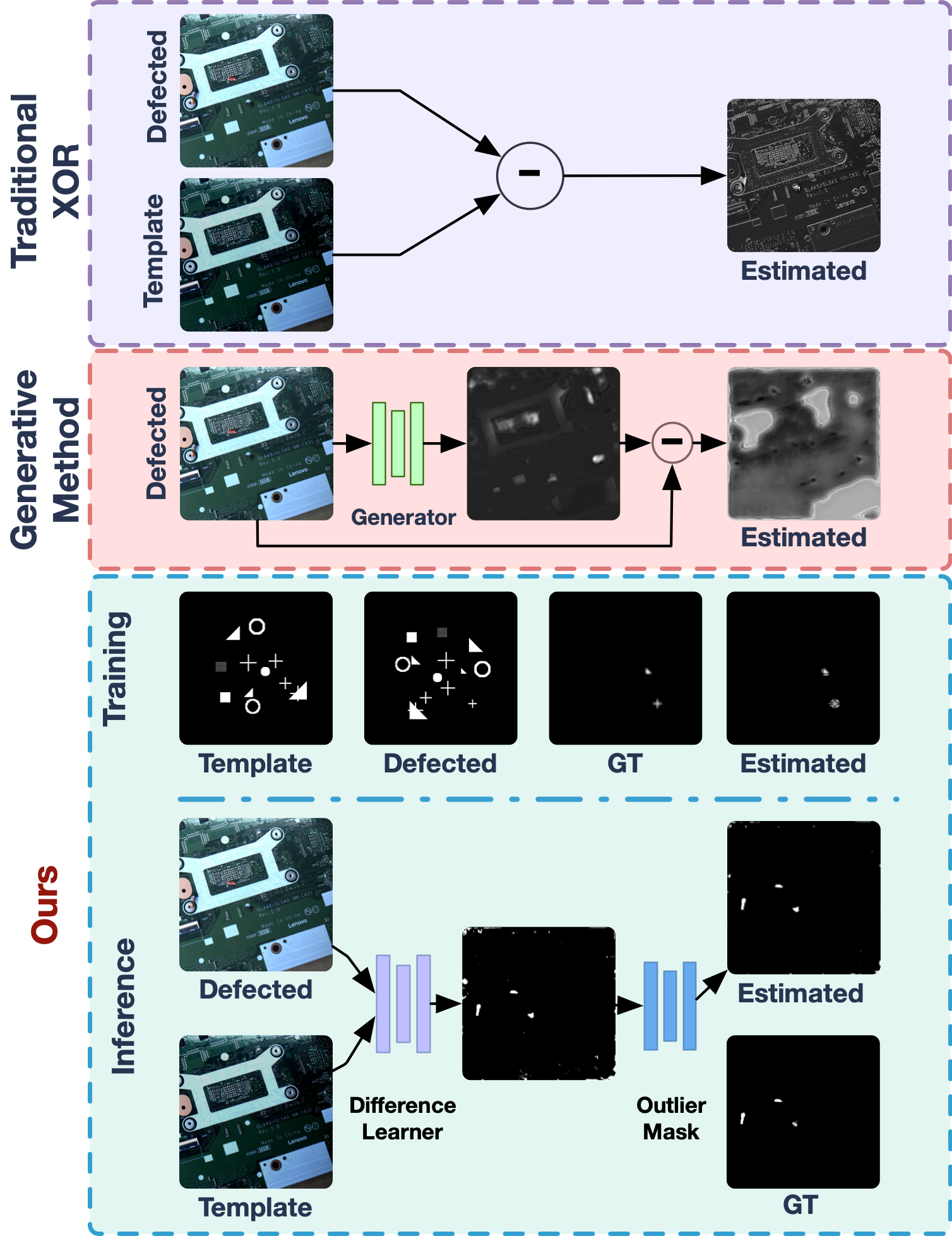}
        \caption{Existing methods for defection segmentation either failed in dealing with lighting changes, e.g. template based XOR logical manipulator (violet box), or lack of generalizing ability, e.g. generative methods (orange box). These make them inapplicable in practical applications where both lighting and context keep changing. Our proposed method (blue box) specifically learns differences between images so that even when trained in simulation, it can still be applied in the real world.}
        \label{fig:teaser}
    \end{figure}
    
Considering one specific circumstance in a printed circuit board (PCB) workshop, where machines are supposed to figuring out random defections in each PCB comparing to its template. Such procedure is challenging since: i) condition changes in each observation leads to the inconvenience in direct comparing, e.g. lighting changes and subtle pose changes between PCBs; ii) pixel-wise error leads to false small defection segmentation, e.g. non-overlapping edges due to assembly error of electronic components in each PCB might be recognized as small defections; iii) PCBs come with different style leads to either the overloaded data and training or the difficulty in the generalization of most of the existing defection segmentation networks; iv) insufficient real-world training data for DS leads to a hard practical implementation. With these difficulties in mind, an ideal approach for PCB defection segmentation should require less expensive real-world training data, and should work with multiple styles of PCBs in a variety of conditions. This task on PCB is representative of many tasks of defection segmentation on random objects, e.g. fingerprint detection in screen manufacturing. The question we ask in this work is: with limited real-world training data and varying PCB styles, how can a network stay functioning.

We answer this question in a sim2real fashion. Instead of training a network to recognize defections in each specific PCB context, we want to endow the robot with the universal ability to distinguish only defections regardless of contextual variance via simulated training. This ability enables the method to generalize to practical applications and eases the burden associated with providing the network with comprehensive data and excessive training, as they are inexhaustible.

To tackle this challenge, we propose a context insensitive small defection segmentation network which is trained on the simulated environment once for all, named \textit{\textit{SSDS}} for ``Sim2Real Small Defection Segmentation Network''. \textit{\textit{SSDS}} is designed to recognize foreign objects considering two given images who are allowed to also have shifts in $\mathbb{SIM}(2)$ space, shown in Fig. \ref{fig:teaser}. The defection-free template is then pose transformed with respect to the pose estimator so that it is pose-aligned to the defected image. To learn to recognize differences between the two images without being misled by the context, a phase correlation based feature extractor is further designed with the backbone of UNet. Utilizing the fact that when two images without share content are fed into deep phase correlation (DPC) \cite{chen2020deep} to calculate the relative pose, the angle output will give a constant false value at $180\degree$ leading to a strong self-supervision constraint forcing the UNet to extract the defection only. Finally, a masking layer further reduces noise introduced by ``static shifting'' such as assembly error of electronic components in PCB inspection. 

The key contribution of the paper is the introduction of guiding a network to distinguish differences between a pair of images to achieve defection segmentation rather than forcing it to remember the context information in the training data. Specific contributions can be summarized as:
\begin{itemize}[leftmargin=*]
    \item \textit{\textit{SSDS}} is proposed to distinguish small defections between an image pair with minimal training data. We guide the network to specifically learn differences between images by making use of the pose sensitivity of phase correlation so that even when trained with cheap simulated data, it can still be generalized to multiple practical applications.
    \item A new dataset \emph{HiDefection} \cite{HiDefection} containing both generated and actual defections on a PCB board is released for defection segmentation researches and this dataset supports the experiments of the proposed method, which demonstrates superior performance of the generalization ability.
\end{itemize}

\section{Related Works}
\label{sec:RW}
Existing methods for defection detection can be summarized into two fashions: template free and template based. Template free methods usually require one input for detection while template based methods need two: defected image and defection-free template. Each of them has its own share of strength.

\subsection{Template Free Defection Segmentation}
\label{subsec:Template Free Defection segmentation}
Template free methods can be roughly divided into models based on pretrained feature extractor, generative adversarial network (GAN), and convolutional autoencoder.

Methods based on pretrained feature extractors usually utilize widely adopted feature extractors as well as classifiers. Napoletano \textit{et al.} \cite{napoletano2018anomaly} utilized ResNet-18 \cite{he2016deep} trained on ImageNet \cite{krizhevsky2012imagenet} to obtain a clustered feature extractor to recognize defections. Andrews \textit{et al.} \cite{andrews2016transfer} achieved the segmentation with the backbone of VGG \cite{simonyan2014very} and utilized a One-Class-SVM \cite{chen2001one} to recognize the defection. These methods are limited by the classifier and are gradually replaced by the following generative methods.

GAN based and autoencoder based methods are generative with the assumption that defections should not be generated since they are unseen in the training set. Schlegl \textit{et al.} \cite{schlegl2017unsupervised} proposed a GAN based manifold extractor that was trained on defection free images. It learned a representation of the manifold of the clear images and expecting it to recognize and localize the defections in a defected image. Ackay \textit{et al.} \cite{akcay2018ganomaly} proposed ``Ganomaly'' and utilized an autoencoder as the generating part of the GAN following by a decoder to generate a positive image. By comparing the generated image to the input image pixel-wise, defections are expected to be revealed. Bergmann \textit{et al.} \cite{bergmann2018improving} generated positive images with an autoencoder and constructed a specific SSIM loss so that the generative error is related to the pixel-wise similarity. Zhai \textit{et al.} \cite{zhai2016deep} introduced energy to the regularized autoencoder so the higher energy in the generated output represented the defections. Rudolph \textit{et al.} \cite{rudolph2021same} detected anomalies via the usage of likelihoods provided by a normalizing flow on multi-scale image features. Comparing to these generative methods, the proposed \textit{\textit{SSDS}} adopts the generative fashion but differs in the content of the generated output: \textit{\textit{SSDS}} generates the defection directly from the defected and defection-free inputs so that even unseen defections can be recognized.

\subsection{Template Based Defection Segmentation}
\label{subsec:Template Based Defection segmentation}
Different from template free approaches, template based methods usually coupe with tiny, vital but untrained defections which can only be recognized by comparing the defected image and the defection-free template. The origin of this is image subtraction \cite{moganti1995automatic} by XOR logical operator, it is easy to apply but with extremely high requirements for pixel-wise alignment of the images. Considering this, Ding \textit{et al.} \cite{ding2019tdd} with feature matching as an improvement has been proposed since it is more robust with pixel errors. Besides feature matching of two images, Gaidhane \textit{et al.} \cite{gaidhane2018efficient} utilized similarity measurement with symmetric matrix to localize the defection. Although these methods succeeded in detecting tiny untrained defections, they are heavily influenced by the context which means the generalization capability is unsatisfying.

    \begin{figure*}[t]
    \centering
        \includegraphics[width=0.95\linewidth]{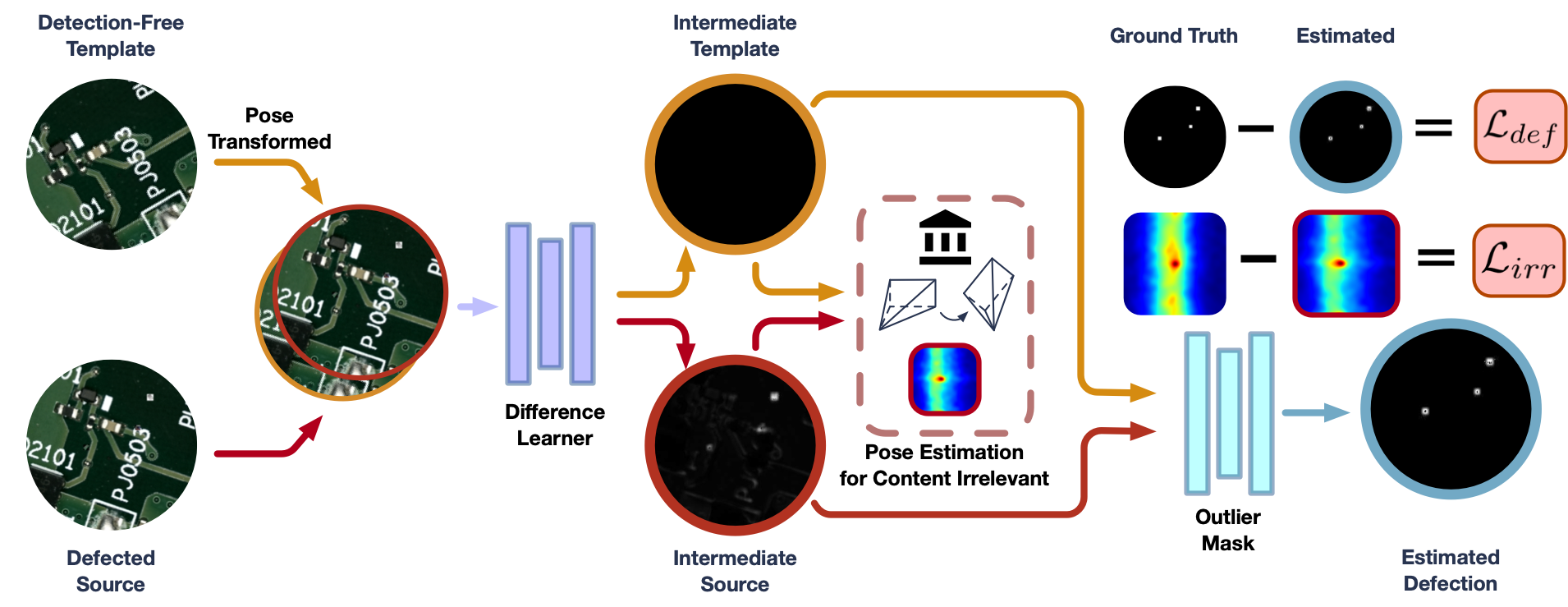}
        \caption{Demonstration of the network structure for \textit{\textit{SSDS}}. Defection-free template is first transformed to be overall pose aligned with the defected source, leaving alone the pixel-wise error caused in practical applications. They are then concatenated to be one single tensor to be processed by the difference learning layer. The pose between intermediate template and source is then calculated and the result of which is further supervised by a ``Irrelevant Loss'' $\mathcal{L}_{irr}$ so that the two intermediate variables' content is different and one of them contains the defections with noises. These intermediate variables are then processed by the outlier masking layer and are further supervised by the ground truth of defection.}
        \label{fig:net_structure}
    \end{figure*}

\textit{\textit{SSDS}} in this paper aims at detecting those untrained, multi-scaled defections with template based inputs but in a generative way. The proposed method is able to generate defection maps between defected and defection-free images regardless of the context so that it can even be trained in simple simulated data to generalize to multiple practical applications where defection-free template is available.

\section{Sim2Real Small Defection Segmentation}
\label{sec:method}

One typical example of the encountered problem is shown in Fig. \ref{fig:teaser} where the defected image and the defection free template are the inputs. Since there are multiple mature methods (mechanically as well as algorithmically \cite{chen2020deep, cheng2019qatm, mughal2021assisting}) that are able to align the images in the industry, we will pass over the pose alignment and assume that the two input images are approximately pose aligned. Following the pipeline in Fig. \ref{fig:net_structure} we design a difference learner for small defection segmentation which aims at only recognizing the differences even if they are small. It is contextual insensitive so that it can be generalized to practical applications with simple training in simulation. Then, a masking module is proposed to eliminate the outlier. Since differentiable pose estimation in $\mathbb{SIM}(2)$ plays an important role in the difference learner, we will introduce it as a prerequisite in Section \ref{subsec:DPC}.

\subsection{Differentiable Phase Correlation}
\label{subsec:DPC}

Given two images $I_{1}$ and $I_{2}$ with pose transformations, a variation of the deep phase correlator \cite{chen2020deep} is utilized to estimate the the overall relative pose $\xi_{i}$ between $I_{1}$ and $I_{2}$. The rotation and scale is calculated by
\begin{gather}
\label{eq: CorrMAP}
p(\xi^{\theta,s}_{i}) = \mathfrak{C}(\mathfrak{L}(\mathfrak{F}(I_{2})),\mathfrak{L}(\mathfrak{F}(I_{1})))\\
(\theta_{i},s_{i}) = \mathop{\mathbb{E}}(p(\xi^{\theta,s}_{i}))
\label{eq: rotation}
\end{gather}
where $\xi^{\theta,s}_{i}$ is the rotation and scale part of $\xi_{i}$ between $I_{1}$, $\mathfrak{F}$ is the discrete Fourier Transform, $\mathfrak{L}$ is the log-polar transform, and $\mathfrak{C}$ is the phase correlation solver. Fourier Transformation $\mathfrak{F}$ transforms images into the Fourier frequency domain of which the magnitude has the property of translational insensitivity, therefore the rotation and scale are decoupled with displacements and are represented in the magnitude. Log-polar transformation $\mathfrak{L}$ transforms Cartesian coordinates into log-polar coordinates so that such rotation and scale in the magnitude of Fourier domain are remapped into displacement in the new coordinates, making it solvable with the afterward phase correlation solver. Phase correlation solver $\mathfrak{C}$ outputs a heatmap indicating the displacements of the two log-polar images, which eventually stands for the rotation  $\theta_{1}$ and scale  $s_{1}$ of the two input images $I_{2}$ and $I_{1}$. To make the solver differentiable, we use expectation $\mathop{\mathbb{E}}(\cdot)$ as the estimation of $\cdot$.

Then $I_{2}$ is rotated and scaled referring to $\xi^{\theta,s}_{i}$ with the result of $\bar{I}_{2}$:
\begin{equation}
 \bar{I}_{2} =  \begin{bmatrix}
s_{i}\boldsymbol{R}_{\theta_{i}} & 0, \\
0 & 1
\end{bmatrix} I_{2}.
\label{eq: Rotate}
\end{equation}

With the same manner, translations $\xi^{\textbf{t}}_{i}$ between $\bar{I}_{2}$ and $I_{1}$ is calculated
\begin{gather}
   p(\xi^{\textbf{t}}_{i}) = \mathfrak{C}(\bar{I}_{2},I_{1}),\\
    \boldsymbol{t}_{i} = \mathop{\mathbb{E}}(p(\xi^{\textbf{t}}_{i})).
    \label{eq: Transform}
\end{gather}


One unique property of deep phase correlation is to identify whether the two input images are related in content. When and only when the two inputs of the phase correlation in the rotation estimation stage are completely irrelevant, the phase correlation map $p(\xi^{\theta,s})$ will be a one-peak distribution centering at exactly $180\degree$ \cite{oppenheim1981importance}. With this property in mind, we will move on to the difference learning module, the difference learning.

\subsection{Difference Learning}
\label{subsec:Learning to Differ}

In this module, we aim at designing a learnable layer $\mathcal{N}_{d}$ that can recognize differences between two input images $I_{s}$ and $I_{t}$ without being disrupted by the context, and this in the practical applications could be defections in PCB, fingerprints in screen manufacturing, etc.
\begin{gather}
    O_{t}, O_{s} = \mathcal{N}_{d}(I_{t}||I_{s}),
    \label{eq: defection learning}
\end{gather}
where $||$ is the concatenation, $O_{t}$ and $O_{s}$ are the desire outputs of $\mathcal{N}_{d}$ with respect to $I_{t}$ and $I_{s}$. The content of $O_{t}$ and $O_{s}$ is further supervised by the proposed ``Irrelevance Loss'' $\mathcal{L}_{irr}$ and by the outlier masking layer $\mathcal{N}_{m}$ introduced in Section \ref{subsec:Outlier Masking}.

The ``Irrelevance Loss'' $\mathcal{L}_{irr}$ forces the content of $O_{t}$ and $O_{s}$ to be completely different by utilizing the similarity identifying ability of deep phase correlation introduced in Section \ref{subsec:DPC}:
\begin{gather}
\label{eq: CorrMAP_irr}
p(\xi^{\theta,s}_{d}) = \mathfrak{C}(\mathfrak{L}(\mathfrak{F}(O_{t})),\mathfrak{L}(\mathfrak{F}(O_{s}))),
\end{gather}
where $p(\xi^{\theta,s}_{d})$ is the phase correlation result on rotation and scale with respect to $O_{t}$ and $O_{s}$. By expecting $O_{t}$ and $O_{s}$ to have completetly irrelevant content, $p(\xi^{\theta,s}_{d})$ is then supervised by the Kullback–Leibler Divergence loss $KLD$ between the Gaussian Blurred one-peak distribution centering at $180\degree$:
\begin{equation}\label{eq: kldxir}
\mathcal{L}_{irr} = KLD(p(\xi^{\theta,s}_{d}),\textbf{1}_{180\degree}).
\end{equation}

By now the content of $O_{t}$ and $O_{s}$ is expected to be completely different from each other. Resulting from this, there are two possibilities for $O_{t}$ and $O_{s}$: i) they are totally in a mess; ii) one of them contains useful information of differences between $I_{t}$ and $I_{s}$. This leads to the next module ``Outlier Masking'' which also constrains the content of $O_{t}$ and $O_{s}$ to be useful.

\subsection{Outlier Masking}
\label{subsec:Outlier Masking}
 Inspired by the problem in Section \ref{subsec:Learning to Differ}, in this module, we define the content of $O_{t}$ and $O_{s}$. We start with focusing on the final defection segmentation result $O$ who is derived from the $O_{t}$ and $O_{s}$ with a learnable masking layer $\mathcal{N}_{m}$:
 \begin{equation}\label{eq: beforemask}
O = \mathcal{N}_{m}(O_{t}||O_{s}).
\end{equation}

The result $O$ should be exactly the same as the ground truth of defection In the coordinate system of $I_{s}$, shown as ``Defection Estimated'' and ``Defection GT'' in Fig. \ref{fig:teaser} respectively. Therefore, a hard supervision based on Mean Square Error $MSE$ on $O$ and the ground truth $G$ is applied:
\begin{equation}
\mathcal{L}_{def} =  {\lVert O-G \lVert_{2}}^{2},
\end{equation}
where ${\lVert \cdot \lVert_{2}}^{2}$ is the $MSE$ loss of $\cdot$. The final loss $\mathcal{L}$ should be:
\begin{equation}\label{eq: loss}
\mathcal{L} = \mathcal{L}_{def} + \mathcal{L}_{irr}.
\end{equation}

Since a masking layer $\mathcal{N}_{m}$ cannot generate defections out of nowhere, at least one of $O_{t}$ and $O_{s}$ should contain the defection features, However, with $\mathcal{L}_{irr}$ introduced (in Section \ref{subsec:Learning to Differ}), $O_{t}$ and $O_{s}$ must not contain same objects so that the defection features should only be presented in one of them, then we have:
\begin{equation}\label{eq: similar}
\exists ! ~ X \in \{O_{t},O_{s}\} (X\sim O),
\end{equation}
which stands for: there exists exactly one variable in the set $\{O_{t},O_{s}\}$ that is similar to the output $O$.

To this end, it is clear that one of $O_{t}$ and $O_{s}$ is a noisy defection map, but we are not concerned about who exactly it is since they are both intermediate results to be processed by $\mathcal{N}_{m}$. This explains that the difference learner learns $\mathcal{N}_{d}$ differences between the two input images and that the context is neglected. With the masking layer proposed, the content of $O_{t}$ and $O_{s}$ is defined and the segmentation is completed.

\subsection{Training in Simulation}
\label{subsec:Training Detail}

\begin{figure}[t]
        \includegraphics[width=\linewidth]{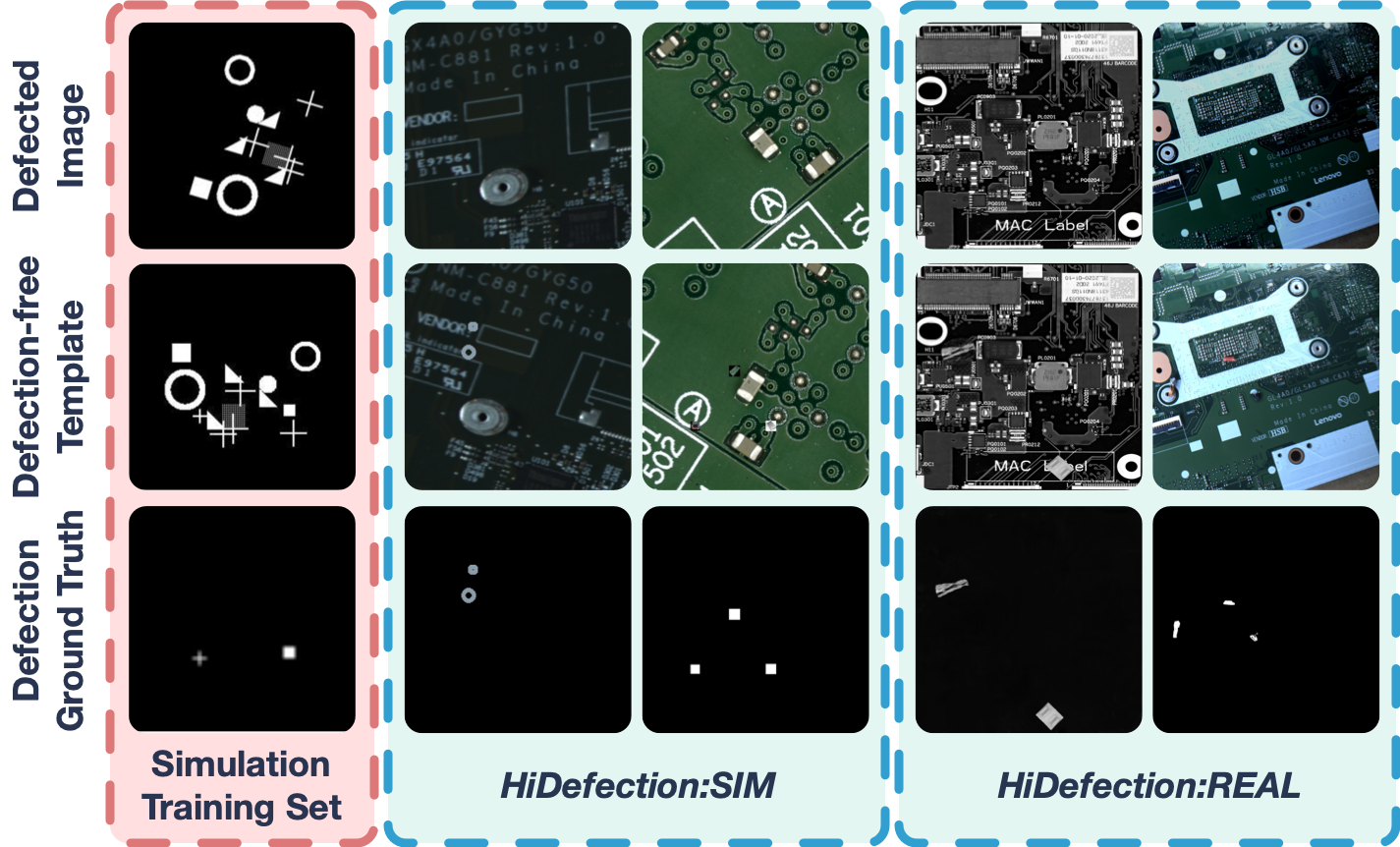}
        \caption{Demonstrations of the simulated training data and \textit{\textbf{HiDefection}} dataset on which the methods are evaluated.}
        \label{fig:dataset}
    \end{figure}

The training skills also play important roles in the methodology since we are designing a once-for-all network needless of further training. One simulated dataset is specifically designed for the \textit{SSDS}'s training with randomization, shown in Fig. \ref{fig:dataset} (Simulation Training Set). One defection-free template is firstly generated with a random number of random shapes at random locations. For the defected image, comparing to the defection-free one, the same number of same shapes are generated around the same location with slight random translational disturbance for imitating practical usage, e.g. assembly error of electronic components. Then the generated defected image is rotated and scaled randomly to simulate the overall pose misalignment in practical applications. The simulated data is completely randomized so that the network will not overfit and will focus on the difference learning. Note that the shapes are partially borrowed from \cite{Unet-pytorch}.


\section{Experiments: Dataset And Setup}
\label{sec:setup}
Our method is trained in simulation (Fig. \ref{fig:dataset}: Simulation Training Set) and evaluated on the sim\&real dataset \textit{HiDefection} \cite{HiDefection} for PCB defection segmentation. As one of the contributions of this paper, \textit{HiDefection} contains two packages \textit{HiDefection:SIM} and \textit{HiDefection:REAL}. \textit{HiDefection:SIM} is constructed with randomly generated shapes as defections on two different types of PCBs as background and \textit{HiDefection:REAL} with realistic photos of actual objects on PCBs.

\subsection{Dataset}
\label{subsec:Dataset}
\noindent \textbf{\textsc{\textit{HiDefection:}}} The \textit{HiDefection} is a real-world dataset collected both in a PCB workshop with simulated defections in different types of PCB and actual defected boards together with their corresponding defection-free templates.
\begin{itemize}[leftmargin=*]
    \item \textbf{\textit{HiDefection:SIM}}: It is recorded in the overhead perspective of a large amount of PCBs and is leveraged in the paper. The random defections in this part of dataset contain random shapes and randomly dropped electronic components. Defections are small with the largest being $0.6\%$ of the whole image. The total size of this part of dataset is 15000 pairs for training and 3000 pairs for validation. One sample pair is shown in Fig. \ref{fig:dataset} as ``\textit{HiDefection:SIM}''.
    \item \textbf{\textit{HiDefection:REAL}}: This sub-dataset is recorded in the same camera condition with \textit{HiDefection:SIM} above but with real defections and PCBs shot together so that it can represent what defection segmentation actually encounters in real-world applications. When collecting, we also introduce light changes in some pairs of images to make it more realistic so that the result on this can be representative of what one method will achieve in practice. Moreover, the assembly error of electronic components are introduced to every pair of data which makes the small defection segmentation more challenging. Note that the size of the largest defection in this part of data will not exceed $1\%$ of the whole image. This part of dataset is designed for validating the generalizing ability of defection segmentation methods and contains 100 pairs of images. One sample pair is shown in Fig. \ref{fig:dataset} as ``\textit{HiDefection:REAL}''.
\end{itemize}

\begin{table}[]
\centering
\caption{Quantitative result for experiments on \textit{HiDefection:SIM} dataset. AP and MaxF1 are calculated between baseline output and the defection ground truth. Note that all baselines are evaluated in \textit{HiDefection:SIM} dataset but trained in either \textit{HiDefection:SIM} or simulation.}
\label{table:resultSIM}
\resizebox{\linewidth}{!}{
\renewcommand{\arraystretch}{1.5}
\begin{tabular}{clllll}
\toprule
\multirow{2}{*}{Scene} &
  \multicolumn{1}{c}{\multirow{2}{*}{\textbf{Baselines}}} &
  \multicolumn{2}{c}{\tabincell{c}{Trained in \\ \textit{HiDefection:SIM}}} &
  \multicolumn{2}{c}{\tabincell{c}{Trained in \\ Simulation}} \\ \cline{3-6} 
 &
  \multicolumn{1}{c}{} &
  \multicolumn{1}{c}{AP($\%$)} &
  \multicolumn{1}{c}{MaxF1($\%$)} &
  \multicolumn{1}{c}{AP($\%$)} &
  \multicolumn{1}{c}{MaxF1($\%$)} \\ \cline{1-2}
\multirow{5}{*}{I}  & \textsc{GANomaly} & 98.2 & 97.5 & 45.4 & 43.1 \\
                    & \textsc{SSIM}     & 97.2 & \textcolor{blue}{98.7} & 38.7 & 39.4 \\
                    & \textsc{DSEBM}    & 98.1 & 94.3 & 28.9 & 42.5 \\
                    & \textsc{TDD-net}  & \textcolor{blue}{98.4} & 98.6 & 50.3 & 67.1 \\
                    & \textsc{\textit{SSDS}}    & $\setminus$ & $\setminus$ & \textcolor{red}{97.4} & \textcolor{red}{95.5} \\ \hline
\multirow{5}{*}{II} & \textsc{GANomaly} & 96.3 & 95.2 & 22.4 & 27.3 \\
                    & \textsc{SSIM}     & 97.0 & 93.4 & 40.3 & 29.9 \\
                    & \textsc{DSEBM}    & \textcolor{blue}{98.4} & 97.0 & 30.5 & 45.2 \\
                    & \textsc{TDD-net}  & 95.6 & \textcolor{blue}{97.7} & 42.9 & 33.5 \\
                    & \textsc{\textit{SSDS}}    & $\setminus$ & $\setminus$ & \textcolor{red}{97.8} & \textcolor{red}{98.2} \\ \bottomrule
\end{tabular}%
}
\end{table}

For all dataset above, defected images are pose transformed from the defection-free template with the constraint on translations of both $x$ and $y$, rotation changes and scale changes of the two images in the range of $[-50, 50]$ pixels, $[0, \pi)$ and $[0.8, 1.2]$ respectively with images shapes of $256 \times 256$.

\subsection{Metrics}
\label{subsec:Metrics}

We evaluate the performance of the defection segmentation with Intersection-over-Union(IoU) AP, Recall Rate R and the Harmony Mean(MaxF1) of AP and R. In Section \ref{sec:results}, AP and MaxF1 are adopted for quantitative demonstration.

\subsection{Comparative Methods}
\label{subsec:Comparative}
Baselines in the experiment include template free methods \textsc{GANomaly}\cite{akcay2018ganomaly}, \textsc{SSIM} \cite{bergmann2018improving}, \textsc{DSEBM}\cite{zhai2016deep} and template based method
\textsc{TDD-net} \cite{ding2019tdd}. All the comparative methods share the same training condition and are trained in the simulated dataset and on \textit{HiDefection}. They are evaluated on \textit{HiDefection} as comparisons to prove the generalization capability of \textit{SSDS}.

\begin{table}[]
\centering
\caption{Quantitative result for experiments on \textit{HiDefection:REAL} dataset. AP and MaxF1 are calculated between baseline outputs and the defection ground truth. Note that all baselines are evaluated in \textit{HiDefection:REAL} dataset but trained in either \textit{HiDefection:SIM} or simulation.}
\label{table:resultREAL}
\resizebox{\linewidth}{!}{
\renewcommand{\arraystretch}{1.5}
\begin{tabular}{clllll}
\toprule
\multirow{2}{*}{Scene} &
  \multicolumn{1}{c}{\multirow{2}{*}{\textbf{Baselines}}} &
  \multicolumn{2}{c}{\tabincell{c}{Trained in \\ \textit{HiDefection:SIM}}} &
  \multicolumn{2}{c}{\tabincell{c}{Trained in \\ Simulation}} \\ \cline{3-6} 
 &
  \multicolumn{1}{c}{} &
  \multicolumn{1}{c}{AP($\%$)} &
  \multicolumn{1}{c}{MaxF1($\%$)} &
  \multicolumn{1}{c}{AP($\%$)} &
  \multicolumn{1}{c}{MaxF1($\%$)} \\ \cline{1-2}
\multirow{5}{*}{III}  & \textsc{GANomaly} & \textcolor{blue}{19.9} & 23.1 & 21.0 & 18.7 \\
                    & \textsc{SSIM}     & 11.1 & \textcolor{blue}{29.3} & 12.9 & 21.1 \\
                    & \textsc{DSEBM}    & 11.9 & 26.2 & 20.5 & 17.5 \\
                    & \textsc{TDD-net}  & 19.2 & 23.8 & 21.4 & 21.9 \\
                    & \textsc{\textit{SSDS}}    & $\setminus$ & $\setminus$ & \textcolor{red}{98.1} & \textcolor{red}{96.3} \\ \hline
\multirow{5}{*}{IV} & \textsc{GANomaly} & 10.6 & 17.1 & 11.0 & 16.9 \\
                    & \textsc{SSIM}     & 13.5 & 12.9 & 16.2 & 20.1 \\
                    & \textsc{DSEBM}    & 26.7 & \textcolor{blue}{21.5} & 39.2 & 43.1 \\
                    & \textsc{TDD-net}  & \textcolor{blue}{33.6} & 19.2 & 29.0 & 31.5 \\
                    & \textsc{\textit{SSDS}}    & $\setminus$ & $\setminus$ & \textcolor{red}{97.1} & \textcolor{red}{94.7} \\ \bottomrule
\end{tabular}%
}
\end{table}

\begin{figure*}[t]
\centering
    \includegraphics[width=0.9\linewidth]{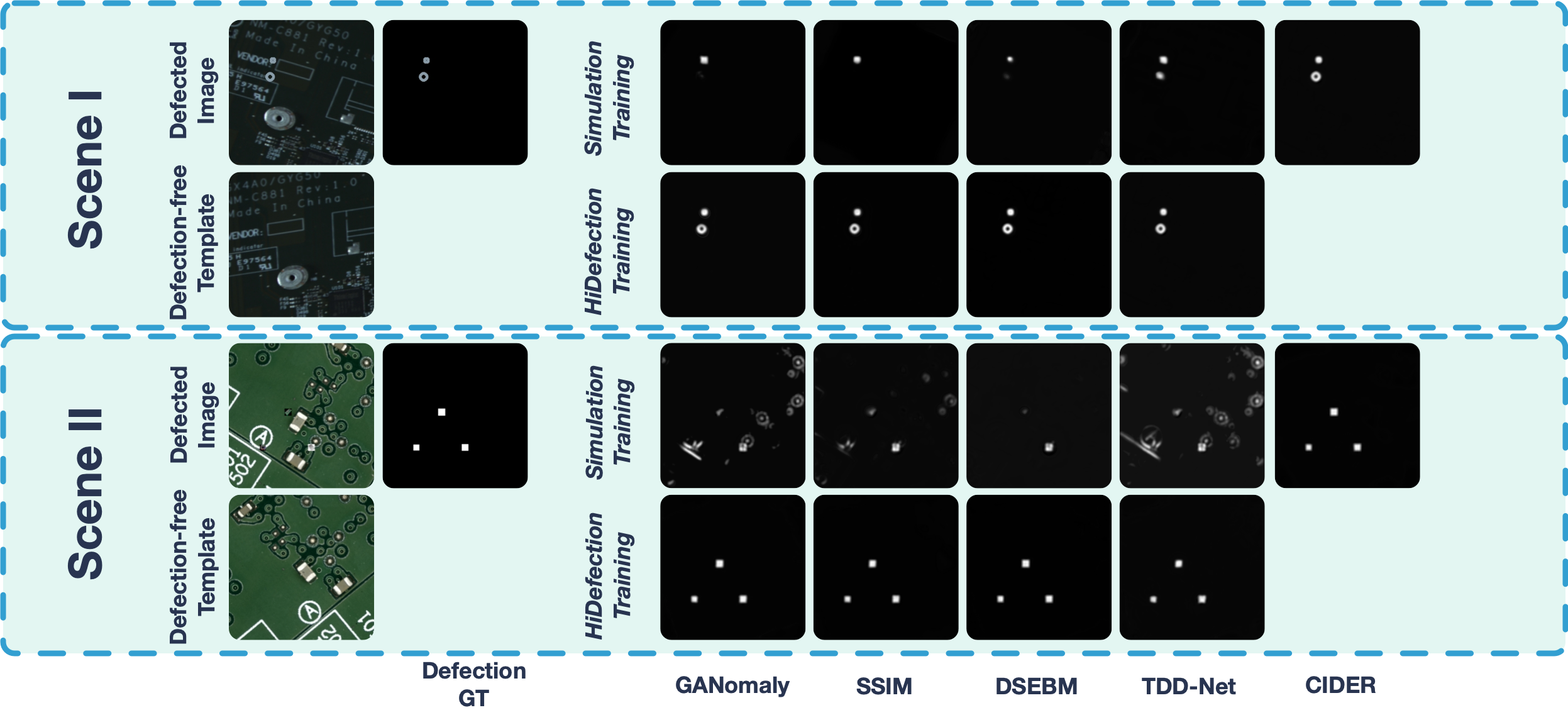}
    \caption{Qualitative comparison of PCB defection segmentation of baselines trained in simulation and \textit{HiDefection:SIM}. Note that \textit{SSDS} is only trained in simulation.}
    \label{fig:resultSIM}
\end{figure*}

\begin{figure*}[t]
\centering
    \includegraphics[width=0.9\linewidth]{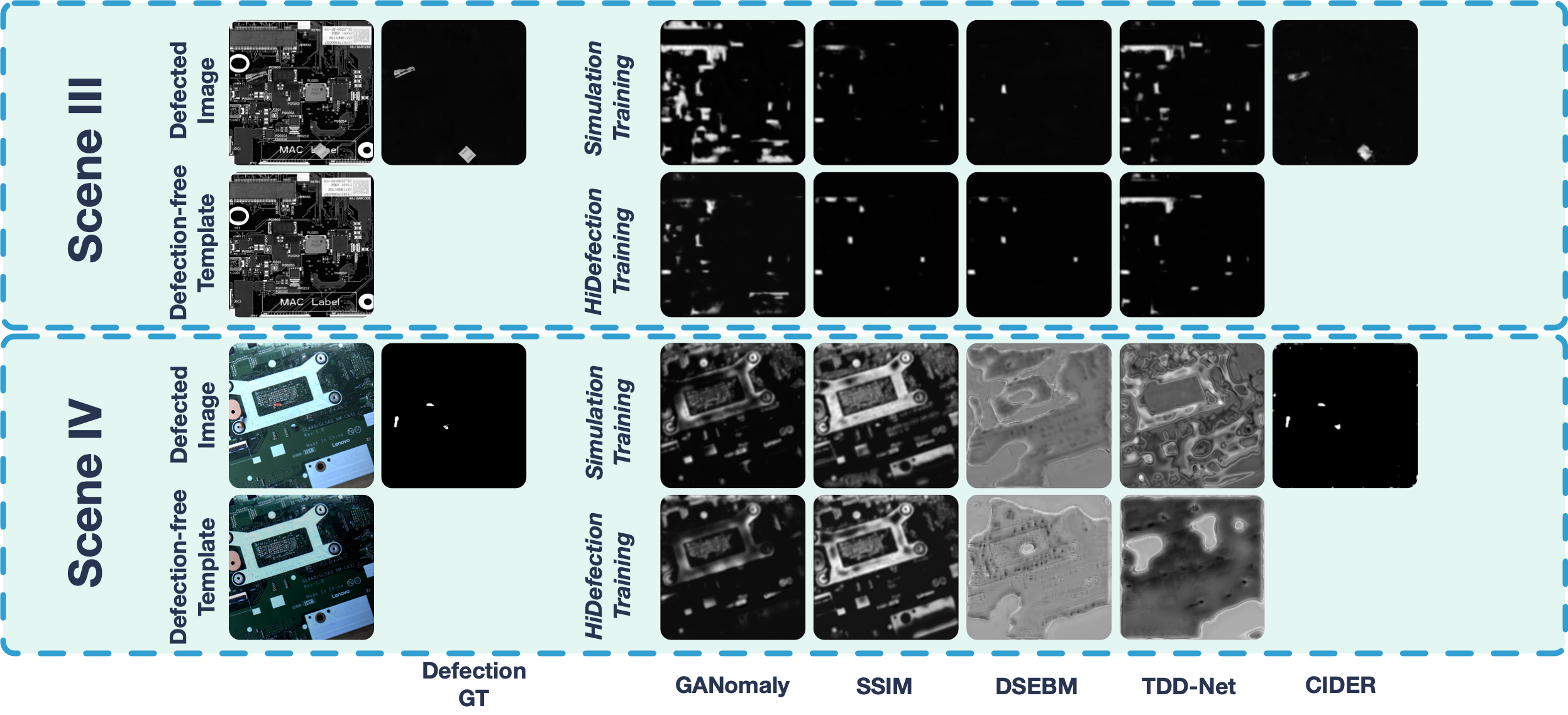}
    \caption{Qualitative comparison of dynamic obstacle detection of baselines trained in simulation and \textit{HiDefection:REAL}. Note that \textit{SSDS} is only trained in simulation.}
    \label{fig:resultREAL}
\end{figure*}

\section{Experiments: Results}
\label{sec:results}

\subsection{PCB with Simulated Defection}
\label{subsec:SIM}

In this experiment, we evaluate \textit{SSDS} on \textit{HiDetection:SIM} where ``Scene I'' has simulated shapes on PCBs and ``Scene II'' has randomly placed electronic components on PCBs. We train \textit{SSDS} and the rest of the baselines in simulation and test it on \textit{HiDetection:SIM}. However, to prove the generalization capability, we additionally train all baselines in \textit{HiDetection:SIM} except for \textit{SSDS}. The qualitative result shown in Fig. \ref{fig:resultSIM} and the quantitative result in TABLE. \ref{table:resultSIM} demonstrate the superior performance \textit{SSDS} achieved in generalizing simulation training to practical inference and it outperforms other baselines in generalizing. Even when trained in simulation with a completely different context, \textit{SSDS} could still be on par with baselines that are trained on \textit{HiDetection:SIM} when comparing the performance on inferring defections of PCBs. When inferred in a different context from training, template free and generative baselines (\textsc{GANomaly, SSIM, DSEBM}) are easily confused by the new context and inevitably miss important defections, and template based baseline \textsc{TDD-net} is misled by the assembly error of the components so that its output is blurry.

In the \textit{HiDetection:SIM} where defections are simulated and are similar to what they are in the simulated training, baselines' generalizing ability is still worthy of segmentation. However, when defections are from the real world, their performances start to drop, and this leads to Section \ref{subsec:REAL}.

\subsection{PCB with Real Defection}
\label{subsec:REAL}
In this experiment, we evaluate \textit{SSDS} on \textit{HiDefection:REAL} where ``Scene III'' and ``Scene IV'' are both real-world particles and PCBs with lighting changes. Different from what has been conducted in Section \ref{subsec:SIM}, all methods including \textit{SSDS} have not been trained on \textit{HiDefection:REAL} so that the generalization is further studied. The qualitative result is shown in Fig. \ref{fig:resultREAL} and the quantitative result is shown in TABLE. \ref{table:resultREAL}. The result indicates that with defections from the real world involved which have not been seen in any of the training data, both the comparing methods trained in simulation and trained in \textit{HiDefection:SIM} fail to present successful predictions of the defection. In contrast, \textit{SSDS} trained in simulation is still a competent approach to recognize defection, even when light changes are involved in one pair of images. This proves the generalizing ability of \textit{SSDS} and that the method can be directly adopted in practical applications without any further training.

\subsection{Ablation Study}
\label{subsec:ABlation}

    \begin{figure}[t]
    \centering
        \includegraphics[width=0.8\linewidth]{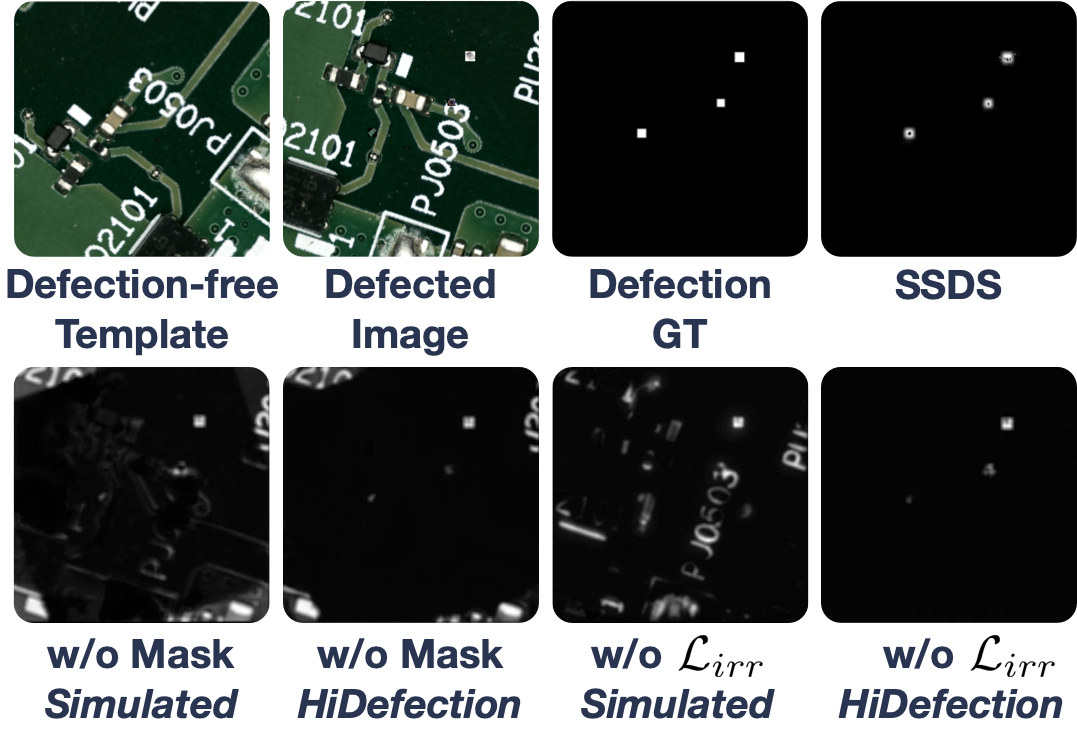}
        \caption{Visual comparison for ablation study. Note that ``w/o $\mathcal{L}_{irr}$ Simulated'' and ``w/o Mask Simulated'' is the specific ablation trained in simulation, ``w/o $\mathcal{L}_{irr}$ \textit{HiDefection}'' and ``w/o Mask \textit{HiDefection}'' are the ablation trained in dataset \textit{HiDefection:SIM}.}
        \label{fig:ablation}
    \end{figure}
    
        \begin{figure}[t]
    \centering
        \includegraphics[width=\linewidth]{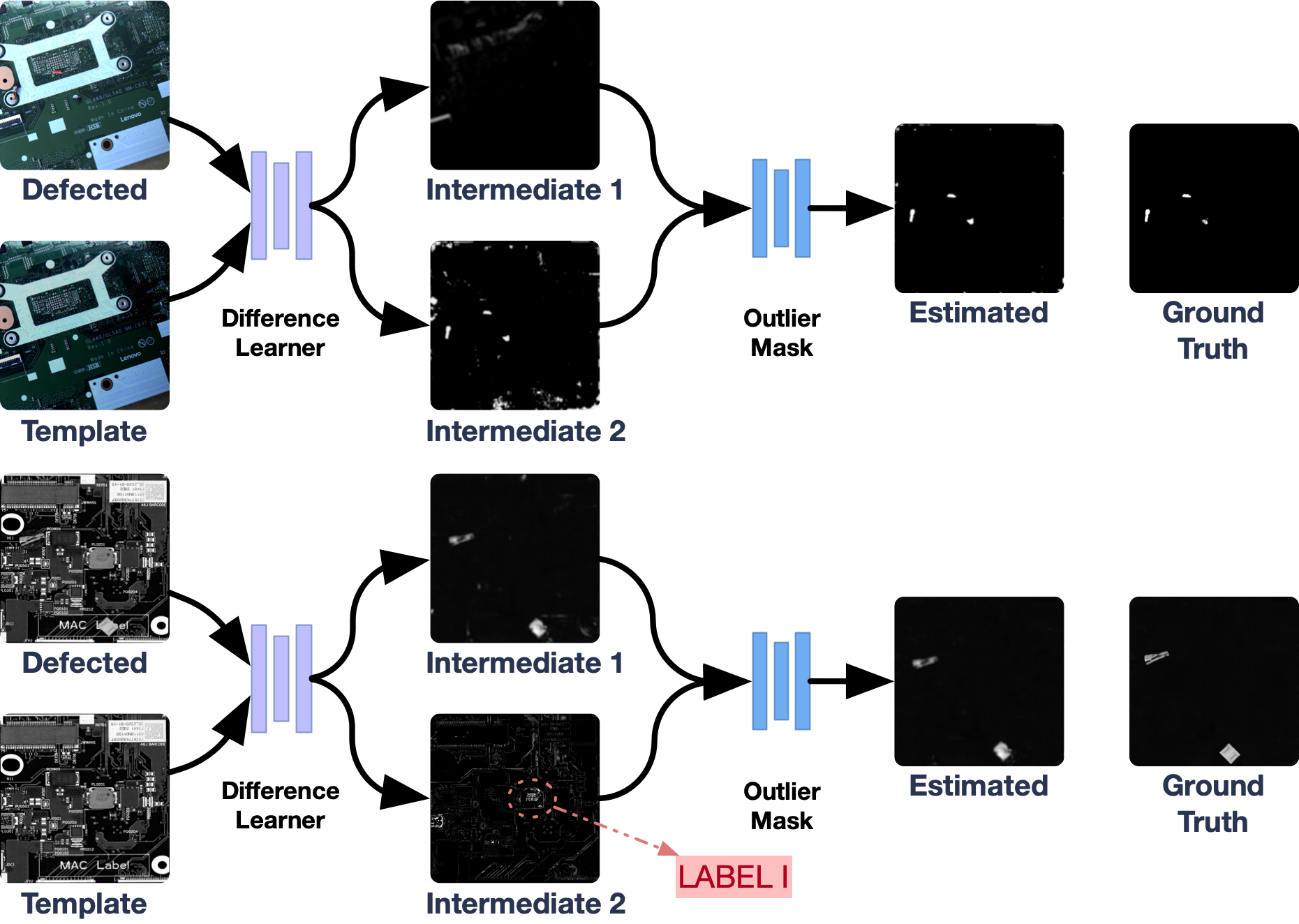}
        \caption{Demonstration of intermediate results in the inferring stage with two independent models trained in simulation to visualize the function of each component of \textit{SSDS}. Note: LABEL I is the difference brought by unshaped silicon grease between two input images and is further removed after the ``Outlier Mask''.}
        \label{fig:ablation_inter}
        \vspace{-10pt}
    \end{figure}
Several experiments are designed for ablation studies to reassure that each module involved plays its own important role. Firstly, we validate the role of the difference learning layer by deprecating this layer and the corresponding loss in the training stage(denotes as ``w/o $\mathcal{L}_{irr}$''). In such study, ``w/o $\mathcal{L}_{irr}$'' is trained both in simulation and on \textit{HiDefection} and evaluated in \textit{HiDefection}. The qualitative result (Fig. \ref{fig:ablation}) and the quantitative result (TABLE. \ref{table:Ablation}) shows that while  ``w/o $\mathcal{L}_{irr}$'' trained on \textit{HiDefection} still gives a satisfying result, it fails in defection segmentation when it is trained in the simulation. This proves that, without the difference learning layer, the method will still work when the training and inferring share the same context, but will fail when they are different. It also proves that the difference learning layer is a vital component for the context insensitive behavior. 

We further study the importance of the masking layer by deprecating such layer and calculate the defection MSE loss $\mathcal{L}_{def}$ directly between $O_{s}$ and $G$ (denotes as ``w/o Mask''). The qualitative result (Fig. \ref{fig:ablation}) and the quantitative result (TABLE. \ref{table:Ablation}) shows that ``w/o Mask'' trained on both \textit{HiDefection} and simulation is able to provide acceptable results in but with more noise generated from the assembly error on components. This proves that the masking layer efficiently eliminates the unfavorable noise due to pixel error.

\subsection{Case Study}
\label{subsec:case study}
We visualize the intermediate results in the whole method to prove our interpretation of the network, shown in Fig. \ref{fig:ablation_inter}. The first case (upper case in Fig. \ref{fig:ablation_inter}) is conducted on a set of PCBs images shot in different lighting conditions, and the second case (lower case in Fig. \ref{fig:ablation_inter}) is where the shapes of silicon grease varies from each other but should not be counted as defection. We trained the network twice in simulation resulting in two independent models. The first model have the ``Intermediate 2'' generated with defections while the second model have it in ``Intermediate 1''. It proves that one of the two outputs of the ``Difference Learner'' contains specific defections with noise introduced by pixel-wise error of the two images and the ``Outlier Mask'' is able to filter the noise out. 

While it is satisfying to prove that \textit{SSDS} is able to deal with lighting changes, its potential of handling images with false defections is rewarding. Note that in the second case, when defections are presented in ``Intermediate 1'', the differences in silicon grease is presented in ``Intermediate 2'' and is filtered out after the ``Outlier Mask''.

    \begin{table}[t]
    \centering
    \textcolor{black}{\caption{Quantitative comparison for ablation studies on \textit{HiDefection:SIM} dataset. It indicates that the main contributing function is the ``Difference Learner" without whom the performance on generalization drops immediately.}
    \label{table:Ablation}
    \resizebox{\linewidth}{!}{%
    \renewcommand{\arraystretch}{1.5}
    \begin{tabular}{@{}lll|lll@{}}
    \toprule
          \multicolumn{1}{c}{\multirow{2}{*}{\textbf{Ablation}}} &
      \multicolumn{2}{c|}{\tabincell{c}{Trained in \\ \textit{HiDefection:SIM}}} &
      \multicolumn{3}{c}{\tabincell{c}{Trained in \\ Simulation}} \\  \cline{2-6}
      \multicolumn{1}{c}{} &
      \multicolumn{1}{c}{w/o $\mathcal{L}_{irr}$} &
      w/o Mask &
      \multicolumn{1}{c}{w/o $\mathcal{L}_{irr}$} &
      w/o Mask &
      \textit{SSDS} \\ \cline{1-1}
    \textsc{AP($\%$)}    & 59.1 &  77.8 &  17.9  & 72.5 & \textcolor{red}{97.8}\\
    \textsc{MaxF1($\%$)} & 63.4 &  74.2 &  19.3  & 78.1 & \textcolor{red}{98.2}\\ \bottomrule
    \end{tabular}%
    }}
    \vspace{-10pt}
    \end{table}

\section{Conclusion}
\label{sec:conclusion}
We present an approach for small defection segmentation with simple training in simulated environment, namely \textit{SSDS}. To achieve this, we design one difference learner based on the deep phase correlation so that it can recognize defections regardless of the context, and one masking layer for outlier elimination. In various experiments, the proposed \textit{SSDS} presents satisfying performances in generalization, i.e. background changes and lighting changes, showing its potential for practical defection segmentation application.

\clearpage

\printbibliography

@article{chen2020deep,
  title={Deep Phase Correlation for End-to-End Heterogeneous Sensor Measurements Matching},
  author={Chen, Zexi and Xu, Xuecheng and Wang, Yue and Xiong, Rong},
  journal={arXiv preprint arXiv:2008.09474},
  year={2020}
}

@inproceedings{schlegl2017unsupervised,
  title={Unsupervised anomaly detection with generative adversarial networks to guide marker discovery},
  author={Schlegl, Thomas and Seeb{\"o}ck, Philipp and Waldstein, Sebastian M and Schmidt-Erfurth, Ursula and Langs, Georg},
  booktitle={International conference on information processing in medical imaging},
  pages={146--157},
  year={2017},
  organization={Springer}
}

@inproceedings{akcay2018ganomaly,
  title={Ganomaly: Semi-supervised anomaly detection via adversarial training},
  author={Akcay, Samet and Atapour-Abarghouei, Amir and Breckon, Toby P},
  booktitle={Asian conference on computer vision},
  pages={622--637},
  year={2018},
  organization={Springer}
}

@article{bergmann2018improving,
  title={Improving unsupervised defect segmentation by applying structural similarity to autoencoders},
  author={Bergmann, Paul and L{\"o}we, Sindy and Fauser, Michael and Sattlegger, David and Steger, Carsten},
  journal={arXiv preprint arXiv:1807.02011},
  year={2018}
}

@inproceedings{zhai2016deep,
  title={Deep structured energy based models for anomaly detection},
  author={Zhai, Shuangfei and Cheng, Yu and Lu, Weining and Zhang, Zhongfei},
  booktitle={International Conference on Machine Learning},
  pages={1100--1109},
  year={2016},
  organization={PMLR}
}

@inproceedings{rudolph2021same,
  title={Same same but differnet: Semi-supervised defect detection with normalizing flows},
  author={Rudolph, Marco and Wandt, Bastian and Rosenhahn, Bodo},
  booktitle={Proceedings of the IEEE/CVF Winter Conference on Applications of Computer Vision},
  pages={1907--1916},
  year={2021}
}

@inproceedings{andrews2016transfer,
  title={Transfer representation-learning for anomaly detection},
  author={Andrews, Jerone and Tanay, Thomas and Morton, Edward J and Griffin, Lewis D},
  year={2016},
  organization={JMLR}
}

@article{napoletano2018anomaly,
  title={Anomaly detection in nanofibrous materials by CNN-based self-similarity},
  author={Napoletano, Paolo and Piccoli, Flavio and Schettini, Raimondo},
  journal={Sensors},
  volume={18},
  number={1},
  pages={209},
  year={2018},
  publisher={Multidisciplinary Digital Publishing Institute}
}

@inproceedings{he2016deep,
  title={Deep residual learning for image recognition},
  author={He, Kaiming and Zhang, Xiangyu and Ren, Shaoqing and Sun, Jian},
  booktitle={Proceedings of the IEEE conference on computer vision and pattern recognition},
  pages={770--778},
  year={2016}
}

@article{krizhevsky2012imagenet,
  title={Imagenet classification with deep convolutional neural networks},
  author={Krizhevsky, Alex and Sutskever, Ilya and Hinton, Geoffrey E},
  journal={Advances in neural information processing systems},
  volume={25},
  pages={1097--1105},
  year={2012}
}

@article{simonyan2014very,
  title={Very deep convolutional networks for large-scale image recognition},
  author={Simonyan, Karen and Zisserman, Andrew},
  journal={arXiv preprint arXiv:1409.1556},
  year={2014}
}

@inproceedings{chen2001one,
  title={One-class SVM for learning in image retrieval},
  author={Chen, Yunqiang and Zhou, Xiang Sean and Huang, Thomas S},
  booktitle={Proceedings 2001 International Conference on Image Processing (Cat. No. 01CH37205)},
  volume={1},
  pages={34--37},
  year={2001},
  organization={IEEE}
}

@article{moganti1995automatic,
  title={Automatic PCB inspection systems},
  author={Moganti, Madhav and Ercal, Fikret},
  journal={IEEE Potentials},
  volume={14},
  number={3},
  pages={6--10},
  year={1995},
  publisher={IEEE}
}

@article{ding2019tdd,
  title={TDD-net: a tiny defect detection network for printed circuit boards},
  author={Ding, Runwei and Dai, Linhui and Li, Guangpeng and Liu, Hong},
  journal={CAAI Transactions on Intelligence Technology},
  volume={4},
  number={2},
  pages={110--116},
  year={2019},
  publisher={IET}
}

@article{gaidhane2018efficient,
  title={An efficient similarity measure approach for PCB surface defect detection},
  author={Gaidhane, Vilas H and Hote, Yogesh V and Singh, Vijander},
  journal={Pattern Analysis and Applications},
  volume={21},
  number={1},
  pages={277--289},
  year={2018},
  publisher={Springer}
}

@online{Unet-pytorch,
    author = "Anonymous",
    url  = "https://github.com/usuyama/pytorch-unet"
}

@online{HiDefection,
    author = "ZJU-Robotics-Lab",
    title = "HiDefection Dataset",
    url  = "https://github.com/ZJU-Robotics-Lab/OpenDataSet",
    year = {2020}
}

@article{mughal2021assisting,
  title={Assisting UAV Localization via Deep Contextual Image Matching},
  author={Mughal, Muhammad Hamza and Khokhar, Muhammad Jawad and Shahzad, Muhammad},
  journal={IEEE Journal of Selected Topics in Applied Earth Observations and Remote Sensing},
  year={2021},
  publisher={IEEE}
}

@inproceedings{cheng2019qatm,
  title={QATM: Quality-aware template matching for deep learning},
  author={Cheng, Jiaxin and Wu, Yue and AbdAlmageed, Wael and Natarajan, Premkumar},
  booktitle={Proceedings of the IEEE/CVF Conference on Computer Vision and Pattern Recognition},
  pages={11553--11562},
  year={2019}
}

@article{wang2016framework,
  title={A framework for multi-session RGBD SLAM in low dynamic workspace environment},
  author={Wang, Yue and Huang, Shoudong and Xiong, Rong and Wu, Jun},
  journal={CAAI Transactions on Intelligence Technology},
  volume={1},
  number={1},
  pages={90--103},
  year={2016},
  publisher={Elsevier}
}

@article{oppenheim1981importance,
  title={The importance of phase in signals},
  author={Oppenheim, Alan V and Lim, Jae S},
  journal={Proceedings of the IEEE},
  volume={69},
  number={5},
  pages={529--541},
  year={1981},
  publisher={IEEE}
}

\end{document}